\documentclass{article}
\usepackage{spconf,amsmath,amssymb,graphicx}

\usepackage{multirow}
\usepackage{comment}
\newcommand{\ind}{\perp\!\!\!\!\perp} 
\usepackage{color, colortbl}
\usepackage{CJKutf8}

\title{Alternate Intermediate Conditioning with Syllable-level and Character-level Targets for Japanese ASR}
\name{Yusuke Fujita, Tatsuya Komatsu, Yusuke Kida}
\address{LINE Corporation, Tokyo, Japan}
\copyrightnotice{978-1-6654-7189-3/22/\$31.00~\copyright2023 IEEE}

\begin{document}
%
\maketitle
\begin{abstract}
End-to-end automatic speech recognition directly maps input speech to characters.
However, the mapping can be problematic when several different pronunciations should be mapped into one character or when one pronunciation is shared among many different characters.
Japanese ASR suffers the most from such many-to-one and one-to-many mapping problems due to Japanese kanji characters.
To alleviate the problems, we introduce explicit interaction between characters and syllables using Self-conditioned connectionist temporal classification (CTC), in which the upper layers are ``self-conditioned'' on the intermediate predictions from the lower layers.
The proposed method utilizes character-level and syllable-level intermediate predictions as conditioning features to deal with mutual dependency between characters and syllables.
Experimental results on Corpus of Spontaneous Japanese show that the proposed method outperformed the conventional multi-task and Self-conditioned CTC methods.
\end{abstract}
\section{Introduction}
End-to-end automatic speech recognition (ASR) simplifies the conventional ASR pipelines by mapping input speech into a text sequence directly, without using a hand-crafted pronunciation dictionary.
Attention-based encoder-decoders \cite{Chorowski15_NIPS} and recurrent neural network transducers \cite{Graves12_ICMLRLW} are popular end-to-end ASR methods in which an autoregressive decoder predicts the next token given previously estimated tokens with an encoded sequence of audio features.
These autoregressive ASR methods have shown state-of-the-art performance with strong encoders such as Transformer \cite{Zhang20_ICASSP} and Conformer \cite{gulati20_interspeech}.

Non-autoregressive ASR methods also have a lot of attention due to their efficient inference, which can predict all tokens simultaneously.
Connectionist temporal classification (CTC) \cite{Graves06_icml} is a fundamental approach to non-autoregressive ASR.
Despite the fact that CTC assumes conditional independence between output labels, which was considered to be a drawback, various approaches on top of CTC \cite{Chen21_SPL,Higuchi20b_interspeech,Chi21_NAACL,lee21_icassp, nozaki21_interspeech,Higuchi21_asru} have shown comparable performance with autoregressive models.
Recently, self-supervised learning models such as wav2vec 2.0 \cite{Baevski20_Neurips} and HuBERT \cite{Hsu21_TASLP} have shown further improvement by pretraining the encoder using a large amount of unlabeled data and finetuning with labeled data with CTC.

Although end-to-end ASR methods achieved sufficient performance in English, the challenge remains in languages such as Japanese, which face a large character vocabulary size with many homophones and multiple pronunciations \cite{Ito17_APSIPA}.
The character vocabulary size is larger than that of phonogram languages such as English.
Japanese has over three thousand characters, while English has at most about one hundred characters.
Japanese ASR also suffers from homophones: many characters share the same pronunciation, e.g., \begin{CJK}{UTF8}{ipxm}高\end{CJK}, \begin{CJK}{UTF8}{ipxm}公\end{CJK}, \begin{CJK}{UTF8}{ipxm}行\end{CJK} and other hundreds of characters have the same pronunciation ``kou''.
Therefore, an acoustic feature should be mapped to different character labels considering language contexts.
In addition, because most Japanese kanji characters have multiple pronunciations, e.g., the character \begin{CJK}{UTF8}{ipxm}空\end{CJK} can be pronounced as ``sora'', ``kara'', ``kuu'', ``a'', ``su'' or others, multiple different acoustic features should be mapped to one character label.
Because of such one-to-many/many-to-one mappings, end-to-end modeling of all these characters becomes problematic due to data scarcity.

For end-to-end modeling under the data scarcity situation, hierarchical multi-task learning with low-level auxiliary targets has been studied \cite{Toshniwal17_interspeech, Rao17_icassp, Sanabria18_slt}.
Loss functions for low-level auxiliary targets, i.e., phonemes or syllables, placed at intermediate layers can help regularize model training for the high-level targets, i.e., characters.
However, there is no interaction between the low-level and high-level targets, although the low-level predictions could narrow down the candidates of the high-level predictions. In addition, if intermediate high-level predictions were given, the low-level predictions could be enhanced.

In this paper, we propose a multi-task learning-based ASR method with explicit interactions between the low-level and high-level targets.
Our study is inspired by Self-conditioned CTC \cite{nozaki21_interspeech}, which estimates intermediate CTC predictions at the lower encoder layer and feeds them back to the upper encoder layers: the upper-layer encoder is ``self-conditioned'' on the lower-layer predictions.
The proposed method utilizes both character-level and syllable-level intermediate predictions as conditioning features fed into the next layers.
The two-level conditioning layers can be placed alternately to deal with {\it mutual} dependency; not only the syllable-to-character predictions but also the character-to-syllable predictions are considered in the proposed method.

Experimental results on Corpus of Spontaneous Japanese (CSJ) \cite{Maekawa2003} show that the proposed alternate intermediate conditioning outperformed conventional multi-task learning-based and Self-conditioned CTC-based methods.
The results also show that the proposed method further reduces syllable error rates than the conventional multi-task learning-based method. The accurate syllable information is particularly useful for downstream tasks of Japanese ASR, such as repeating a spoken query by the system with the correct pronunciation.

We also experiment with Mandarin and English, which also face the homophone and multiple pronunciation problems, not as much as Japanese.
Experimental results with Mandarin AISHELL-1 \cite{Bu17_ococosda} and LibriSpeech-100 \cite{Librispeech} show that the proposed alternate intermediate conditioning also outperformed the conventional methods.

\section{Related Work}

\subsection{CTC-based non-autoregressive ASR}
Non-autoregressive ASR methods developed on top of CTC can be categorized into iterative refinement decoding \cite{Chen21_SPL, Higuchi20b_interspeech, Chi21_NAACL} and intermediate prediction objectives \cite{lee21_icassp, nozaki21_interspeech}.

Iterative refinement decoding relaxes the conditional independence assumption between output labels by using a non-autoregressive refinement decoder on top of the CTC-based encoder.
The decoder is trained with the masked language model objective function of token sequences \cite{Higuchi20b_interspeech} or alignment paths \cite{Chi21_NAACL}. The decoder iteratively refines an initial prediction from the encoder by editing low-confident characters with the masked language model.
A drawback of the approach is that the iterative refinement processes increase the computational complexity.

In contrast, our proposed method does not use a decoder while adopting intermediate prediction objectives to the CTC-based encoder.
Intermediate CTC \cite{lee21_icassp} uses auxiliary CTC losses at the intermediate encoder layers, which helps faster optimization to the CTC task in lower layers.
Self-conditioned CTC \cite{nozaki21_interspeech} further utilizes the intermediate CTC predictions to enhance the encoder by feeding the intermediate predictions to the next encoder layer.
We extend the Self-conditioned CTC approach by utilizing syllable-level auxiliary targets as well as character-level targets.

\subsection{Multi-task ASR with low-level auxiliary targets}

Multi-task learning with low-level tasks has been actively studied \cite{Toshniwal17_interspeech, Rao17_icassp, Sanabria18_slt}.
In typical hierarchical multi-task models, low-level auxiliary tasks are placed at intermediate layers, while the main task is always placed only at the highest layer.
Our proposed method differs from these methods in that the predictions for the main task are also applied in intermediate layers, and both low-level and high-level intermediate predictions are utilized for conditioning the subsequent layers.

The grapheme and phoneme sequences can be simultaneously produced by a single-sequence one-to-one model \cite{Omachi21_naacl} that combines the multiple sequences into a single sequence.
This approach is attractive because no architectural modification from typical end-to-end ASR models is required.
However, it was difficult to outperform a separate grapheme-only model, possibly due to increased data scarcity.

Joint phoneme-grapheme model \cite{Kubo20_icassp} utilizes both phone-level and grapheme-level intermediate predictions with iterative refinement decoding. This model can handle mutual dependency between phoneme and grapheme sequences, while it requires two autoregressive decoders.
Aiming to capture the mutual dependency, we propose a simpler non-autoregressive model without using autoregressive decoders or iterative refinement decoding.

Our study is inspired by HC-CTC \cite{Higuchi22_icassp}, which extends Self-conditioned CTC by using multiple training targets.
The multi-granular sub-word units are placed hierarchically according to the vocabulary sizes.
Since their targets are character-based sub-words, it is not directly applicable to ideogram languages with large character vocabulary sizes.
In this paper, we propose to use syllable-level targets instead of sub-word-level targets and examine non-hierarchical placements of the intermediate targets.

\section{Self-conditioned CTC}

In this section, we introduce Self-conditioned CTC \cite{nozaki21_interspeech} to describe necessary notations for the proposed method.

\subsection{Conformer-CTC}
\label{sec:conformerctc}
End-to-end ASR models the posterior distribution of a $L$-length character sequence $Y=(y_l \in \mathcal{V} \mid l=1,\dots,L)$ given a $T$-length input sequence $X=(\mathbf{x}_t \in \mathbb{R}^D \mid t=1,\dots,T)$, where $\mathcal{V}$ is a character vocabulary and $D$ is the dimension of acoustic features.
CTC \cite{Graves06_icml} handles the posterior $p(Y|X)$ via frame-level alignment paths between $X$ and $Y$ with a $\mathsf{blank}$ symbol.
The path is denoted by $A=\left(a_t \in \mathcal{V}' \mid t=1,\dots,T\right)$, where $\mathcal{V}'= \mathcal{V} \cup \{\mathsf{blank}\}$.
The path can be mapped to the character sequence by using the collapsing function $\mathcal{B}$ that removes all repeated characters and blank symbols.
CTC trains a neural network that predicts the path.
An output sequence of the neural network is denoted by $Z = (\mathbf{z}_t \in (0,1)^{|\mathcal{V}'|} \mid t=1,\dots,T)$, where the element $z_{t,k}$ is interpreted as the posterior probability of the character $k$ at time $t$: $p(a_t = k |X)$.
The neural network is trained by minimizing the following CTC loss between an output sequence $Z$ and the character sequence $Y$:
\begin{align}\label{eq:lossctc}
    \mathcal{L}_\mathsf{ctc}(Z, Y) = - \log \sum_{A\in\mathcal{B}^{-1}(Y)}\prod_{t} z_{t,a_t},
\end{align}
which is the negative log-likelihood over possible paths with the conditional independence assumption: $(a_t \ind a_{\ne t} \mid X)$.

The neural network used in this paper has $N$ Conformer encoders \cite{gulati20_interspeech} that accept an input sequence $X^{(n-1)}$ and outputs the encoded sequence:
\begin{align}
    \label{eq:enc}
    X^{(n)} = \mathsf{Encoder}^{(n)}(X^{(n-1)}) \qquad (1 \le n \le N),
\end{align}
where $X^{(0)} = X$.
The output sequence $Z$ is obtained by applying a linear and the softmax functions to the encoded sequence:
\begin{align}
\label{eq:out}
    Z = \mathsf{Softmax}(\mathsf{Linear}_{D\rightarrow |\mathcal{V}'|}(X^{(N)})),
\end{align}
where $\mathsf{Linear}_{D\rightarrow |\mathcal{V}'|}(\cdot)$ maps a $D$-dimensional vector into  a $|\mathcal{V}'|$-dimensional vector for each element in the input sequence.

\subsection{Intermediate CTC}
Intermediate CTC \cite{lee21_icassp} introduces additional CTC predictions from intermediate encoder blocks. An intermediate prediction sequence for the $n$-th encoder block $Z^{(n)} = (\mathbf{z}^{(n)}_t \in (0,1)^{|\mathcal{V}'|}| t=1,\dots,T)$ is computed as:
\begin{align}
\label{eq:softmax}
    Z^{(n)} = \mathsf{Softmax}(\mathsf{Linear}_{D\rightarrow |\mathcal{V}'|}(X^{(n)})).
\end{align}
Note that the linear layer is shared with the output layer (Eq. \ref{eq:out}) so that no additional parameters are added.
The losses for the intermediate predictions are computed as well as the original CTC loss (Eq. \ref{eq:lossctc}), and the total loss is a weighted sum of the original and the intermediate CTC losses:
\begin{equation}
    \mathcal{L}_\mathsf{ic} = (1-\lambda)\mathcal{L}_\mathsf{ctc}(Z,Y) + \frac{\lambda}{|\mathcal{N}|}\sum_{n \in \mathcal{N}} \mathcal{L}_\mathsf{ctc}(Z^{(n)},Y),
\end{equation}
where $\lambda \in (0,1)$ is a mixing weight and $\mathcal{N}$ is a set of layer indices for intermediate loss computation.

\subsection{Self-conditioned CTC}
Self-conditioned CTC \cite{nozaki21_interspeech} further utilizes the intermediate CTC prediction as a condition for the next encoder input. Eq.~\ref{eq:enc} is modified as follows:
\begin{align}
    X^{(n)} &= \mathsf{Encoder}^{(n)}(X'^{(n-1)}) \qquad (1 \le n \le N), \\
    X'^{(n)} &=
        \begin{cases}
            X^{(n)} + \mathsf{Linear}_{|\mathcal{V}'|\rightarrow D}(Z^{(n)}) & (n \in \mathcal{N}), \\
            X^{(n)} & (n \notin \mathcal{N}),
        \end{cases}
\label{eq:selfcond}
\end{align}
where $\mathsf{Linear}_{|\mathcal{V}'|\rightarrow D}(\cdot)$ maps a $|\mathcal{V}'|$-dimensional vector into a $D$-dimensional vector for each element in the input sequence.
This linear layer is shared among the $|\mathcal{N}|$ intermediate layers.
In this way, the upper-layer encoder is ``self-conditioned'' on the lower-layer predictions.
This intermediate conditioning relaxes the conditional independence assumption of CTC-based ASR models since the upper-layer encoder can see intermediate output tokens embedded into the encoder input.

The proposed method is an extension of Self-conditioned CTC. We utilize the intermediate conditioning technique not only with the same character-based CTC objective but together with the low-level auxiliary targets.

\section{Proposed Method: Alternate Intermediate Conditioning}
\begin{figure}[tb]
  \centering
  \includegraphics[width=0.7\linewidth]{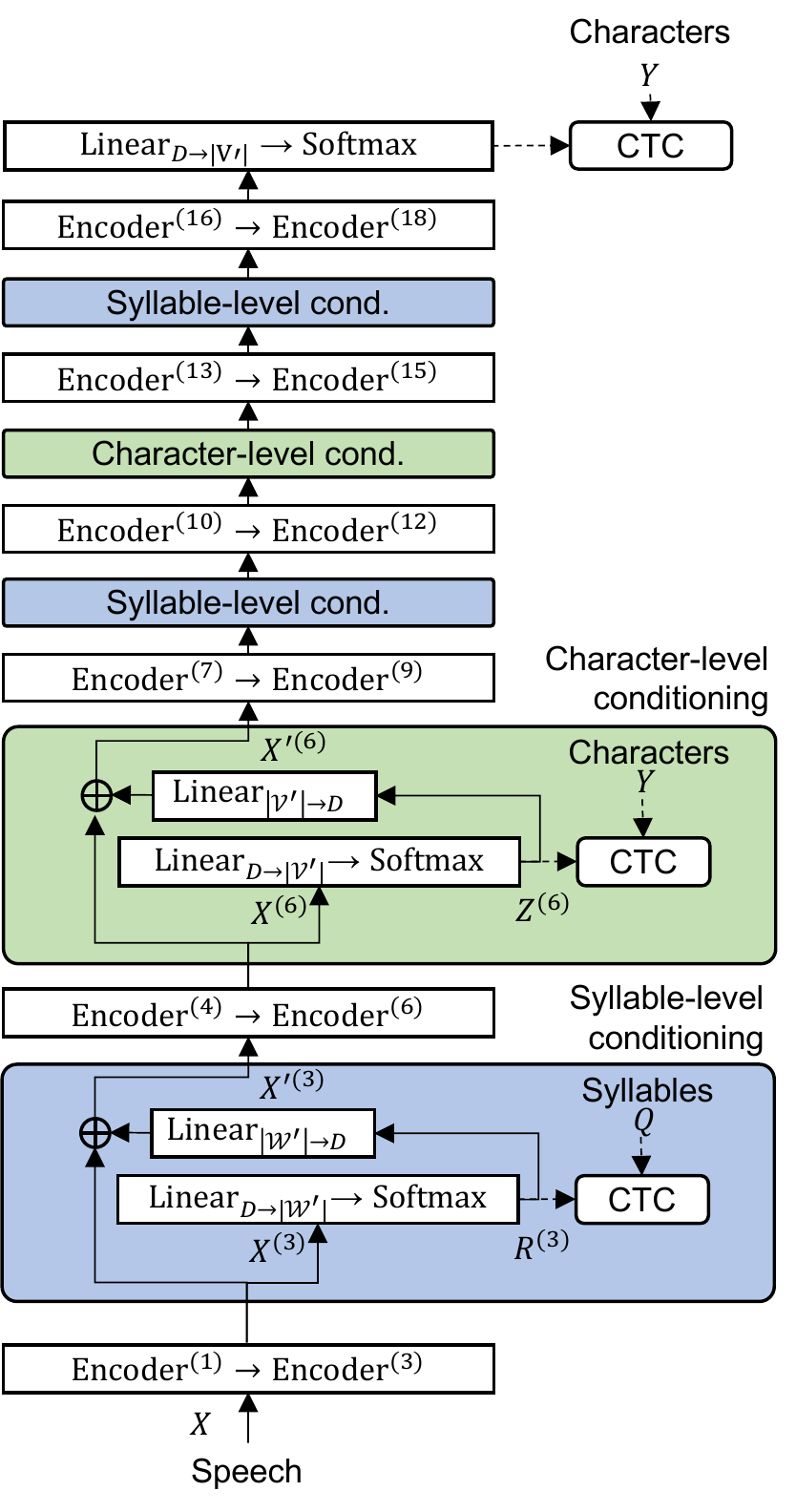}
  \caption{The overview of the proposed alternate intermediate conditioning method. Intermediate conditioning is performed every three encoder blocks and syllable-level conditioning and character-level conditioning are placed alternately.}
  \label{fig:proposed}
\end{figure}
Figure \ref{fig:proposed} shows an overview of the proposed alternate intermediate conditioning method.
A syllable-level auxiliary target sequence  $Q=(q_m \in \mathcal{W} \mid m=1,\dots,M)$ is prepared for each training sample,
where $\mathcal{W}$ is a set of syllables and $M$ is the length of the syllable sequence.
Additional CTC predictions $R^{(n)} = (\mathbf{r}_t \in (0,1)^{|\mathcal{W}'|}|t=0,\dots,T)$ for the syllable sequence are computed from intermediate layers, similar to Eq.~\ref{eq:softmax}:
\begin{equation}
    R^{(n)} = \mathsf{Softmax}(\mathsf{Linear}_{D\rightarrow |\mathcal{W}'|}(X^{(n)})),
\end{equation}
where $\mathcal{W}' = \mathcal{W} \cup \{\mathsf{blank}\}$.
The total loss is a weighted sum of the original CTC loss, the intermediate CTC loss for characters, and the intermediate CTC loss for syllables:
\begin{align}
    \mathcal{L}_\mathsf{mic} &= (1-\lambda)\mathcal{L}_\mathsf{ctc}(Z,Y) \\ \nonumber
    &+ \frac{\lambda}{|\mathcal{N}|+|\mathcal{N}_q|} \sum_{n \in \mathcal{N}} \mathcal{L}_\mathsf{ctc}(Z^{(n)},Y) \\ \nonumber
    &+ \frac{\lambda}{|\mathcal{N}|+|\mathcal{N}_q|} \sum_{n \in \mathcal{N}_q} \mathcal{L}_\mathsf{ctc}(R^{(n)},Q),
\end{align}
where $\mathcal{N}_q$ is a set of layer indices for the intermediate loss for syllables.
As with Self-conditioned CTC, the intermediate predictions are fed back to the next encoder, as follows:
\begin{align}
\label{eq:multiseqcond}
    X'^{(n)}_t &=
        \begin{cases}
            X^{(n)} + \mathsf{Linear}_{|\mathcal{V}'|\rightarrow D}(Z^{(n)}) & (n \in \mathcal{N} \setminus \mathcal{N}_q), \\
            X^{(n)} + \mathsf{Linear}_{|\mathcal{W}'|\rightarrow D}(R^{(n)}) & (n \in \mathcal{N}_q \setminus \mathcal{N}), \\
            X^{(n)} + \mathsf{Linear}_{|\mathcal{V}'|\rightarrow D}(Z^{(n)}) \\
            \qquad \; + \; \mathsf{Linear}_{|\mathcal{W}'|\rightarrow D}(R^{(n)}) & (n \in \mathcal{N} \cap \mathcal{N}_q), \\
            X^{(n)} & (n \notin \mathcal{N} \cup \mathcal{N}_q). \\
        \end{cases}
\end{align}

The sets of intermediate layers $\mathcal{N}$ and $\mathcal{N}_q$ are important hyperparameters in the proposed method. In Figure \ref{fig:proposed}, an example of the sets of layers for the proposed ``alternate'' intermediate conditioning is shown: $\mathcal{N} = \{6,12\}$ and $\mathcal{N}_q = \{3,9,15\}$, where character-level and syllable-level conditioning layers are placed alternately to handle the mutual dependency between the two tasks.

We experiment with ``hierarchical'' intermediate conditioning for comparison: syllable-level predictions are placed at lower layers, while character-level predictions are at higher layers.
We also investigate ``parallel'' intermediate conditioning. 
In this case, syllable-level and character-level conditioning are done at the same layer by adding two embeddings from both $Z^{(n)}$ and $R^{(n)}$ to the next encoder input.

\section{Experiment}

\begin{table*}[t]
\centering
\caption{Character error rates on CSJ. The results were obtained without language models.}
\label{tab:main}
\begin{tabular}{cccccrrr}
\hline
&\multirow{2}{*}{Method} & \multicolumn{2}{c}{\begin{tabular}[c]{@{}c@{}} Layer indices for CTC loss\end{tabular}} & \multirow{2}{*}{\begin{tabular}[c]{@{}l@{}}Intermediate\\ conditioning\end{tabular}} & \multicolumn{3}{c}{CERs (\%)} \\
& & \multicolumn{1}{c}{Character} & \multicolumn{1}{c}{Syllable} & & \multicolumn{1}{c}{eval1} & \multicolumn{1}{c}{eval2} & \multicolumn{1}{c}{eval3} \\ \hline
Conventional&Baseline          & 18 & & N &  5.4 & 3.9 & 9.9  \\
&Multitask        & 18 & 15 & N & 5.4 & 3.7 & 9.5 \\
&InterCTC         & 3,6,9,12,15,18 & & N & 5.4 & 3.8 & 9.4 \\
&SelfCond    & 3,6,9,12,15,18 & & Y & 5.3 & 3.7 & 9.2 \\ \hline
Proposed &Parallel          & 6,12,18 & 6,12,18 & Y & 5.2 & 3.6 & \bf 8.8 \\
&Hierarchical      & 12,15,18 & 3,6,9 & Y & \bf 5.1 & 3.6 & \bf 8.8 \\
&Alternate         & 6,12,18 & 3,9,15 & Y & \bf 5.1 & \bf 3.5 & \bf 8.8 \\ \hline
Prior study&CTC/Att+BeamSearch \cite{Higuchi21_asru}& & & & \bf 5.1 & 3.8 & 9.0 \\
&Self-conditioned CTC \cite{Higuchi21_asru}& & & & 5.3 & 3.7 & 9.1 \\ \hline
\end{tabular}
\end{table*}

\begin{table*}[t]
\centering
\caption{Examples of character-level and syllable-level intermediate predictions with the proposed model with alternate conditioning. The underlined characters indicate substitution errors and \underline{**} indicates deletion errors.}
\label{tab:example}
\begin{tabular}{rll} \hline
Layer & Syl/Chr & Prediction \\ \hline
3 & Syl & \footnotesize{\begin{CJK}{UTF8}{ipxm}ダ イ イ チ \underline{**} レー ノ シ \underline{ケ} ン シュー テ ン ト ダ イ ニ シ レー ノ シ テ ン ガ カ ナ ラ ズ イ \underline{**} \underline{キ} ス ル\end{CJK}} \\
9 & Syl & \footnotesize{\begin{CJK}{UTF8}{ipxm}ダ イ イ チ シ レー ノ シ テ ン シュー テ ン ト ダ イ ニ シ レー ノ シ テ ン ガ カ ナ ラ ズ イ ッ チ ス ル\end{CJK}} \\
15 & Syl & \footnotesize{\begin{CJK}{UTF8}{ipxm}ダ イ イ チ シ レー ノ シ テ ン シュー テ ン ト ダ イ ニ シ レー ノ シ テ ン ガ カ ナ ラ ズ イ ッ チ ス ル\end{CJK}} \\ \hline
6 & Chr & \footnotesize{\begin{CJK}{UTF8}{ipxm}第 一 \underline{**} \underline{例} の \underline{試} \underline{け} 終 点 と 第 二 指 令 の \underline{視} 点 が 必 ず 一 致 す る\end{CJK}} \\
12 & Chr & \footnotesize{\begin{CJK}{UTF8}{ipxm}第 一 \underline{**} 令 の 始 点 終 点 と 第 二 指 令 の 始 点 が 必 ず 一 致 す る
\end{CJK}} \\
18 & Chr & \footnotesize{\begin{CJK}{UTF8}{ipxm}第 一 指 令 の 始 点 終 点 と 第 二 指 令 の 始 点 が 必 ず 一 致 す る
\end{CJK}} \\ \hline
\end{tabular}

\end{table*}

To verify the effectiveness of the proposed method, we conducted experiments conforming to the comparative study for non-autoregressive ASR models \cite{Higuchi21_asru},
by using ESPnet \cite{watanabe18_interspeech, Pengcheng21_icassp} with almost the same hyperparameters.

\subsection{Data}
Our main results were obtained using CSJ \cite{Maekawa2003}.
The 271-hour subset of academic presentation speech (CSJ-APS) was used for training.
Since the corpus includes not only character sequences but also corresponding pronunciation labels, syllable-level sequences were extracted from the pronunciation labels as the auxiliary training targets.
The character vocabulary size $|\mathcal{V}|$ was 2753, and the syllable vocabulary size $|\mathcal{W}|$ was 256.
We used the official evaluation sets: ``eval1'', ``eval2'', and ``eval3'' for testing.
Although CSJ contains another 330-hour subset of simulated public speech on everyday topics (CSJ-SPS) for training, we did not use CSJ-SPS to conform to the prior study \cite{Higuchi21_asru}.
Note that ``eval1'' and ``eval2'' were drawn from CSJ-APS, while ``eval3'' was drawn from CSJ-SPS, which was considered an out-of-domain test set.

Mandarin AISHELL-1 corpus \cite{Bu17_ococosda} was also used for evaluating the proposed method.
According to the official split of AISHELL-1, the 150-hour subset was used for training.
For syllable-level targets, pinyin labels without tones were automatically generated from character sequences by using {\it pypinyin} \footnote{https://github.com/mozillazg/python-pinyin}.
The character vocabulary size $|\mathcal{V}|$ was 4231, and the syllable vocabulary size $|\mathcal{W}|$ was 404.

An additional experiment was conducted on 100-hour subset of LibriSpeech~\cite{Librispeech}.
For the English experiment, 
300 subwords tokenized with SentencePiece~\cite{kudo-richardson-2018-sentencepiece} were used instead of characters as the main output target: $|\mathcal{V}| = 300$. The low-level auxiliary targets are prepared using the CMU pronunciation dictionary\footnote{http://www.speech.cs.cmu.edu/cgi-bin/cmudict}. The phoneme sequence within each word was concatenated and then tokenized into 300 subwords: $|\mathcal{W}| = 300$.
\footnote{The English experiment used the phoneme-based subwords instead of syllables because the number of English syllables was too large to form the output layer. Although the phoneme-based subwords were not actually syllables, they were derived from the phoneme sequences as well as syllables.}

For the input samples, 80-dimensional Mel-scale filterbank coefficients with three-dimensional pitch features were extracted using Kaldi toolkit \cite{Povey11_ASRU}.
Speed perturbation \cite{Ko15_interspeech} and SpecAugment \cite{Park19_interspeech} were also applied to the training data.

\subsection{Model configurations}

We prepared four models with conventional methods and three variant models for our proposed method.
Note that the number of parameters in these models was 32 million for CSJ, 31 million for LibriSpeech-100, and 52 million for AISHELL-1, respectively. The number of additional parameters introduced by the proposed method was small, at most 0.2 million, since the additional linear layers were shared among intermediate layers.

\subsubsection{Conventional models}
\textbf{Baseline:}
The Conformer-CTC model as described in Section \ref{sec:conformerctc} was used.
The number of layers $N$ was 18, and the encoder dimension $D$ was 256.
The convolution kernel size and the number of attention heads were 15 and 4, respectively.
According to the character vocabulary size, the feed-forward layer dimension in the Conformer blocks was set differently: 1024 for CSJ and LibriSpeech-100, and 2048 for AISHELL-1.
The CSJ and LibriSpeech-100 models were trained for 50 epochs, while the AISHELL-1 models were trained for 100 epochs. The final model was obtained by averaging model parameters over 10-best checkpoints in terms of validation loss values.
The Adam optimizer \cite{Kingma14_iclr} with $\beta_1 = 0.9$, $\beta_2=0.98$, the Noam learning rate scheduling \cite{Vaswani17_NIPS} with 25k warmup steps, a learning rate factor of 5.0 were used for training.
For CSJ and AISHELL-1, four GPUs are used in parallel with a batch size of 128 per GPU. For LibriSpeech-100, one GPU is used with a batch size of 128.
CTC greedy decoding \cite{Graves06_icml} was used at inference, without using any language models.

\noindent\textbf{Multitask:}
Based on the Baseline model, an additional syllable-level CTC prediction was placed at the 15-th layer, a few layers lower than the last layer, which is similar to conventional multi-task learning methods \cite{Toshniwal17_interspeech}.

\noindent\textbf{InterCTC:}
Five character-level intermediate CTC predictions were placed at $\mathcal{N} = \{3,6,9,12,15\}$ with $\lambda = 0.5$.
Other configurations were identical to the Baseline model.

\noindent\textbf{SelfCond:}
In addition to the InterCTC model, conditioning with the intermediate CTC predictions was applied.

\subsubsection{Proposed models}
We experimented with three strategies regarding the placement of intermediate predictions: $\mathcal{N}$ and $\mathcal{N}_q$.
As with the SelfCond model, which had five intermediate CTC predictions, we placed the same number of CTC predictions for the proposed model.

\noindent\textbf{Parallel:}
Character-level and syllable-level predictions were placed at the same layers: $\mathcal{N} = \{6,12\}$ and $\mathcal{N}_q = \{6, 12, 18\}$ with $\lambda=0.5$.

\noindent\textbf{Hierarchical:}
Character-level predictions were placed at higher layers $\mathcal{N} = \{12, 15\}$, while syllable-level predictions were placed at lower layers $\mathcal{N}_q = \{3, 6, 9\}$ with $\lambda=0.5$.

\noindent\textbf{Alternate:}
Character-level and syllable-level CTC predictions were placed alternately as shown in Fig. \ref{fig:proposed}: $\mathcal{N} = \{6, 12\}$ and $\mathcal{N}_q = \{3, 9, 15\}$ with $\lambda = 0.5$.

\subsection{Results on CSJ}

Table \ref{tab:main} shows the character error rates for CSJ evaluation sets.
The results obtained from the conventional models show that the multi-task learning model with auxiliary syllable-level predictions performed better than the baseline CTC-based model. However, a similar performance was obtained by Intermediate CTC without using the syllable-level predictions. Self-conditioned CTC was the best model among the conventional methods.

The proposed models outperformed the conventional models. The results clearly show that the syllable-level intermediate predictions help improve the accuracy when they are used together with character-level intermediate predictions.
We could not see a clear difference between the three conditioning strategies, while the alternate conditioning was slightly better than the others.
The results suggest that the alternate conditioning can capture mutual dependency between syllables and characters.
Table \ref{tab:example} shows examples of the intermediate predictions that gradually reduce errors layer by layer.

The proposed models showed significant improvement over conventional models on ``eval3'', which was considered
as an out-of-domain test set.
The results suggest that the auxiliary syllable-level prediction can improve robustness against domain mismatch, which is particularly important for large vocabulary ideogram languages with a relatively small number of training samples per character.

Table \ref{tab:syl} shows syllable error rates on CSJ, which were obtained using the intermediate layer outputs.
The proposed method achieved better accuracy on the low-level auxiliary task than the conventional multi-task learning method.
Syllable error rates were reduced layer by layer.
The results suggest that accurate syllable recognition information from the intermediate layer are fed back to the encoder on the upper layers and this conditioning helps improve the character-level predictions. 

\begin{table}[t]
\centering
\caption{Syllable error rates on CSJ obtained from intermediate predictions.
}
\label{tab:syl}
\begin{tabular}{ccccrrr}
\hline
Method & Layer & eval1 & eval2 & eval3 \\ \hline
Multitask & 15 & 4.0 & 2.6 & 4.8 \\ \hline
Alternate & 3 & 8.4 & 5.4 & 8.8 \\
 & 9 & 4.2 & 2.5 & 4.8 \\
 & 15 & \bf 3.8 & \bf 2.3 & \bf 4.3 \\ \hline
\end{tabular}
\end{table}

\begin{figure}[tb]
  \centering
  \includegraphics[width=0.8\linewidth]{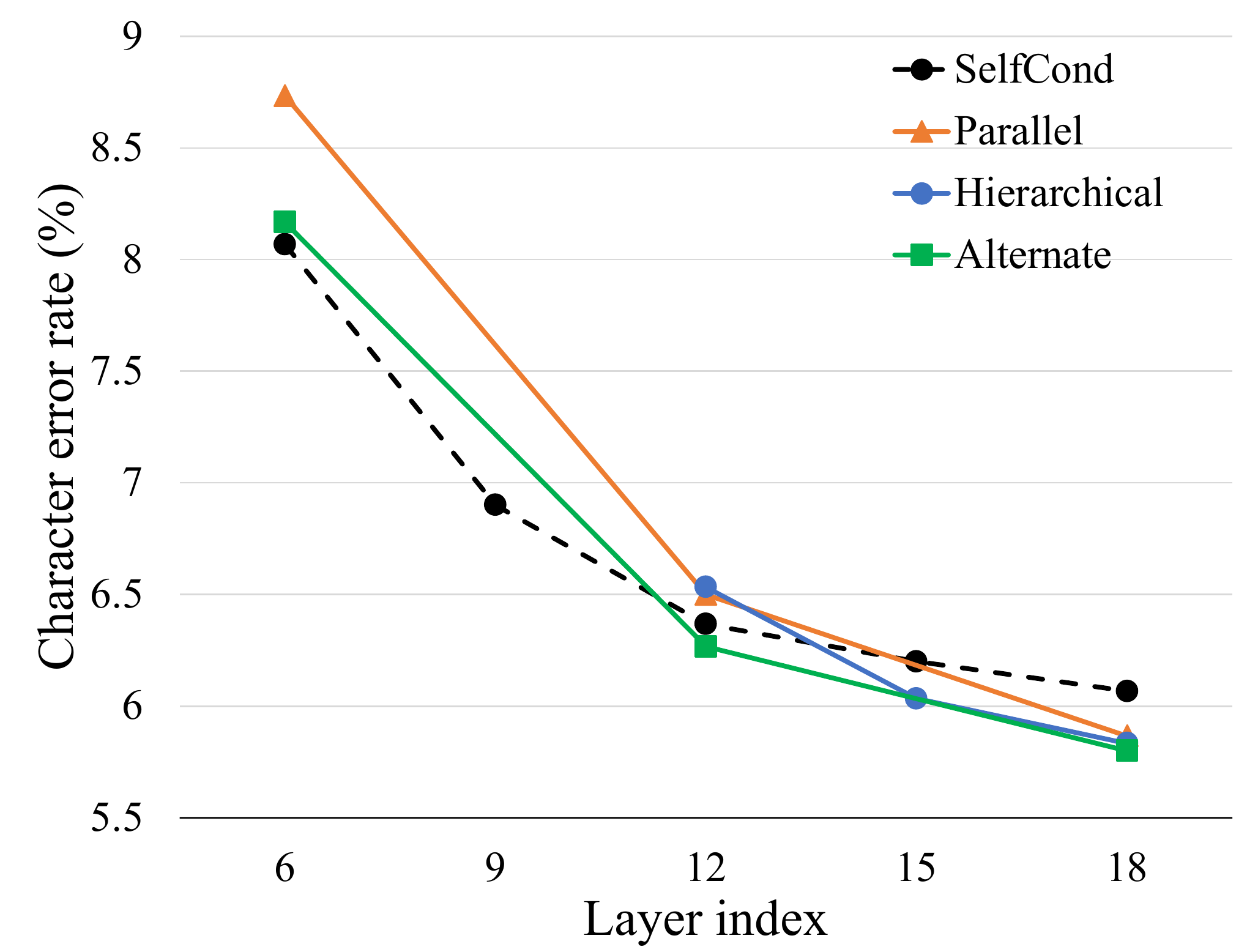}
  \caption{Character error rates on CSJ obtained from intermediate and final predictions. The error rates on three datasets are averaged.}
  \label{fig:intercer}
\end{figure}

Fig.~\ref{fig:intercer} shows character error rates on CSJ obtained from intermediate layers.
Similar to syllable error rates shown in Table \ref{tab:syl}, character error rates were reduced layer by layer.
Conventional self-conditioned CTC achieved slightly better results on lower than 12th layer, while the error reduction at the higher layer was smaller than the proposed method.
At the 12th intermediate layer, we see that alternate intermediate conditioning was better than other strategies. The results suggest that conditioning with two different-level targets alternately gives better intermediate predictions and leads to better final performance.

\subsection{Results on AISHELL-1 and LibriSpeech-100}

Table \ref{tab:aishell} shows the results with Mandarin AISHELL-1.
The proposed method with alternate conditioning strategy was the best among the proposed conditioning strategies.
Contrary to the CSJ results, parallel conditioning showed worse performance than the others.
The results suggest that the two-level conditioning placed at the same layer can hurt performance, while the alternate intermediate conditioning can give consistent performance improvement for both Japanese and Mandarin.
Table \ref{tab:libri100} shows the results with LibriSpeech-100. Consistent performance improvement was observed even with the phonogram language.

\begin{table}[t]
\centering
\caption{Character error rates on AISHELL-1. The results were obtained without language models.}
\label{tab:aishell}
\begin{tabular}{crr}
\hline
Method & dev & test \\ \hline
Baseline & 5.6 & 6.0 \\
Multitask & 4.9 & 5.4 \\
InterCTC & 4.4 & 4.8 \\
SelfCond & 4.2 & 4.6 \\ \hline
Parallel     & 4.5 & 4.9  \\
Hierarchical & 4.1 & 4.5 \\
Alternate    & \bf 4.0 & \bf 4.3 \\ \hline
Conformer-Transformer \cite{Chang21_asru} & 4.4 & 4.7\\
Gated Interlayer Collaboration \cite{YYang22_arxiv}  & \bf 4.0 & 4.4 \\ \hline
\end{tabular}
\end{table}

\begin{table}[t]
\centering
\caption{Word error rates on Librispeech-100. The results were obtained without language models.}
\label{tab:libri100}
\setlength{\tabcolsep}{2.5pt}
\begin{tabular}{crrrr}
\hline
Method & dev-clean & dev-other & test-clean & test-other \\ \hline
SelfCond & 7.1 & 20.8 & 7.6 & 21.1\\ 
Alternate & \bf 6.7 & \bf 19.9 & \bf 7.1 & \bf 20.0 \\ \hline
\end{tabular}
\end{table}

\subsection{Comparison with prior studies}

Comparison results with the prior comparative study \cite{Higuchi21_asru} for CSJ and recently published papers \cite{Chang21_asru, YYang22_arxiv} for AISHELL-1 are shown in Table \ref{tab:main} and \ref{tab:aishell}, respectively.
The proposed model with alternate conditioning outperformed the best non-autoregressive ASR methods and improved variants of Self-conditioned CTC.
From these results, it can be concluded that the proposed method achieves state-of-the-art performance by utilizing syllable-level targets.

\section{Conclusions}

We proposed an end-to-end ASR method to model the interaction between the syllable-level and character-level predictions using intermediate conditioning technique with Self-conditioned CTC.
Experimental results on CSJ show that the proposed method outperformed the conventional multi-task and Self-conditioned CTC methods.

\bibliographystyle{IEEEbib}
\bibliography{mybib}

\begin{thebibliography}{10}

\bibitem{Chorowski15_NIPS}
Jan Chorowski, Dzmitry Bahdanau, Dmitriy Serdyuk, Kyunghyun Cho, and Yoshua
  Bengio,
\newblock ``Attention-based models for speech recognition,''
\newblock in {\em Proc. NIPS}, 2015.

\bibitem{Graves12_ICMLRLW}
Alex Graves,
\newblock ``Sequence transduction with recurrent neural networks,''
\newblock in {\em International Conference on Machine Learning: Representation
  Learning Workshop}, 2012.

\bibitem{Zhang20_ICASSP}
Qian Zhang, Han Lu, Hasim Sak, Anshuman Tripathi, Erik McDermott, Stephen Koo,
  and Shankar Kumar,
\newblock ``Transformer transducer: A streamable speech recognition model with
  transformer encoders and rnn-t loss,''
\newblock in {\em Proc. ICASSP}, 2020, pp. 7829--7833.

\bibitem{gulati20_interspeech}
Anmol Gulati, James Qin, Chung-Cheng Chiu, Niki Parmar, Yu~Zhang, Jiahui Yu,
  Wei Han, Shibo Wang, Zhengdong Zhang, Yonghui Wu, and Ruoming Pang,
\newblock ``{Conformer: Convolution-augmented Transformer for Speech
  Recognition},''
\newblock in {\em Proc. Interspeech}, 2020, pp. 5036--5040.

\bibitem{Graves06_icml}
Alex Graves, Santiago Fern\'{a}ndez, Faustino Gomez, and J\"{u}rgen
  Schmidhuber,
\newblock ``Connectionist temporal classification: Labelling unsegmented
  sequence data with recurrent neural networks,''
\newblock in {\em Proc. ICML}, 2006, p. 369–376.

\bibitem{Chen21_SPL}
Nanxin Chen, Shinji Watanabe, Jesús Villalba, Piotr Żelasko, and Najim Dehak,
\newblock ``{Non-Autoregressive Transformer for Speech Recognition},''
\newblock {\em IEEE Signal Processing Letters}, vol. 28, pp. 121--125, 2021.

\bibitem{Higuchi20b_interspeech}
Yosuke Higuchi, Shinji Watanabe, Nanxin Chen, Tetsuji Ogawa, and Tetsunori
  Kobayashi,
\newblock ``{Mask CTC: Non-Autoregressive End-to-End ASR with CTC and Mask
  Predict},''
\newblock in {\em Proc. Interspeech}, 2020, pp. 3655--3659.

\bibitem{Chi21_NAACL}
Ethan~A. Chi, Julian Salazar, and Katrin Kirchhoff,
\newblock ``{Align-Refine: Non-Autoregressive Speech Recognition via Iterative
  Realignment},''
\newblock in {\em NAACL}, 2021.

\bibitem{lee21_icassp}
Jaesong Lee and Shinji Watanabe,
\newblock ``{Intermediate Loss Regularization for CTC-Based Speech
  Recognition},''
\newblock in {\em Proc. ICASSP}, 2021, pp. 6224--6228.

\bibitem{nozaki21_interspeech}
Jumon Nozaki and Tatsuya Komatsu,
\newblock ``{Relaxing the Conditional Independence Assumption of CTC-Based ASR
  by Conditioning on Intermediate Predictions},''
\newblock in {\em Proc. Interspeech}, 2021, pp. 3735--3739.

\bibitem{Higuchi21_asru}
Yosuke Higuchi, Nanxin Chen, Yuya Fujita, Hirofumi Inaguma, Tatsuya Komatsu,
  Jaesong Lee, Jumon Nozaki, Tianzi Wang, and Shinji Watanabe,
\newblock ``A comparative study on non-autoregressive modelings for
  speech-to-text generation,''
\newblock in {\em Proc. ASRU}, 2021, pp. 47--54.

\bibitem{Baevski20_Neurips}
Alexei Baevski, Yuhao Zhou, Abdelrahman Mohamed, and Michael Auli,
\newblock ``wav2vec 2.0: A framework for self-supervised learning of speech
  representations,''
\newblock in {\em Proc. NeurIPS}, 2020, vol.~33, pp. 12449--12460.

\bibitem{Hsu21_TASLP}
Wei-Ning Hsu, Benjamin Bolte, Yao-Hung~Hubert Tsai, Kushal Lakhotia, Ruslan
  Salakhutdinov, and Abdelrahman Mohamed,
\newblock ``{HuBERT: Self-Supervised Speech Representation Learning by Masked
  Prediction of Hidden Units},''
\newblock {\em IEEE/ACM Transactions on Audio, Speech, and Language
  Processing}, vol. 29, pp. 3451--3460, 2021.

\bibitem{Ito17_APSIPA}
Hitoshi Ito, Aiko Hagiwara, Manon Ichiki, Takeshi Mishima, Shoei Sato, and Akio
  Kobayashi,
\newblock ``End-to-end speech recognition for languages with ideographic
  characters,''
\newblock in {\em 2017 Asia-Pacific Signal and Information Processing
  Association Annual Summit and Conference (APSIPA ASC)}, 2017, pp. 1228--1232.

\bibitem{Toshniwal17_interspeech}
Shubham Toshniwal, Hao Tang, Liang Lu, and Karen Livescu,
\newblock ``{Multitask Learning with Low-Level Auxiliary Tasks for
  Encoder-Decoder Based Speech Recognition},''
\newblock in {\em Proc. Interspeech}, 2017, pp. 3532--3536.

\bibitem{Rao17_icassp}
Kanishka Rao and Haşim Sak,
\newblock ``Multi-accent speech recognition with hierarchical grapheme based
  models,''
\newblock in {\em Proc. ICASSP}, 2017, pp. 4815--4819.

\bibitem{Sanabria18_slt}
Ramon Sanabria and Florian Metze,
\newblock ``{Hierarchical Multitask Learning With CTC},''
\newblock in {\em Proc. SLT}, 2018, pp. 485--490.

\bibitem{Maekawa2003}
Kikuo Maekawa,
\newblock ``Corpus of spontaneous japanese: Its design and evaluation,''
\newblock in {\em ISCA \& IEEE Workshop on Spontaneous Speech Processing and
  Recognition}, 2003.

\bibitem{Bu17_ococosda}
Hui Bu, Jiayu Du, Xingyu Na, Bengu Wu, and Hao Zheng,
\newblock ``{AISHELL-1: An open-source Mandarin speech corpus and a speech
  recognition baseline},''
\newblock in {\em 20th Conference of the Oriental Chapter of the International
  Coordinating Committee on Speech Databases and Speech I/O Systems and
  Assessment (O-COCOSDA)}, 2017, pp. 1--5.

\bibitem{Librispeech}
Vassil Panayotov, Guoguo Chen, Daniel Povey, and Sanjeev Khudanpur,
\newblock ``Librispeech: An asr corpus based on public domain audio books,''
\newblock in {\em 2015 IEEE International Conference on Acoustics, Speech and
  Signal Processing (ICASSP)}, 2015, pp. 5206--5210.

\bibitem{Omachi21_naacl}
Motoi Omachi, Yuya Fujita, Shinji Watanabe, and Matthew Wiesner,
\newblock ``End-to-end {ASR} to jointly predict transcriptions and linguistic
  annotations,''
\newblock in {\em Proc. NAACL-HLT}. June 2021, pp. 1861--1871, Association for
  Computational Linguistics.

\bibitem{Kubo20_icassp}
Yotaro Kubo and Michiel Bacchiani,
\newblock ``Joint phoneme-grapheme model for end-to-end speech recognition,''
\newblock in {\em Proc. ICASSP}, 2020, pp. 6119--6123.

\bibitem{Higuchi22_icassp}
Yosuke Higuchi, Keita Karube, Tetsuji Ogawa, and Tetsunori Kobayashi,
\newblock ``{Hierarchical Conditional End-to-End ASR with CTC and
  Multi-Granular Subword Units},''
\newblock in {\em Proc. ICASSP}, 2021.

\bibitem{watanabe18_interspeech}
Shinji Watanabe, Takaaki Hori, Shigeki Karita, Tomoki Hayashi, Jiro Nishitoba,
  Yuya Unno, Nelson {Enrique Yalta Soplin}, Jahn Heymann, Matthew Wiesner,
  Nanxin Chen, Adithya Renduchintala, and Tsubasa Ochiai,
\newblock ``{ESPnet: End-to-End Speech Processing Toolkit},''
\newblock in {\em Proc. Interspeech}, 2018, pp. 2207--2211.

\bibitem{Pengcheng21_icassp}
Pengcheng Guo, Florian Boyer, Xuankai Chang, Tomoki Hayashi, Yosuke Higuchi,
  Hirofumi Inaguma, Naoyuki Kamo, Chenda Li, Daniel Garcia-Romero, Jiatong Shi,
  Jing Shi, Shinji Watanabe, Kun Wei, Wangyou Zhang, and Yuekai Zhang,
\newblock ``Recent developments on espnet toolkit boosted by conformer,''
\newblock in {\em Proc. ICASSP}, 2021, pp. 5874--5878.

\bibitem{kudo-richardson-2018-sentencepiece}
Taku Kudo and John Richardson,
\newblock ``{S}entence{P}iece: A simple and language independent subword
  tokenizer and detokenizer for neural text processing,''
\newblock in {\em Proceedings of the 2018 Conference on Empirical Methods in
  Natural Language Processing: System Demonstrations}, Brussels, Belgium, Nov.
  2018, pp. 66--71, Association for Computational Linguistics.

\bibitem{Povey11_ASRU}
Daniel Povey, Arnab Ghoshal, Gilles Boulianne, Lukas Burget, Ondrej Glembek,
  Nagendra Goel, Mirko Hannemann, Petr Motlicek, Yanmin Qian, Petr Schwarz, Jan
  Silovsky, Georg Stemmer, and Karel Vesely,
\newblock ``The {Kaldi} speech recognition toolkit,''
\newblock in {\em Proc. ASRU}, 2011.

\bibitem{Ko15_interspeech}
Tom Ko, Vijayaditya Peddinti, Daniel Povey, and Sanjeev Khudanpur,
\newblock ``{Audio augmentation for speech recognition},''
\newblock in {\em Proc. Interspeech}, 2015, pp. 3586--3589.

\bibitem{Park19_interspeech}
Daniel~S. Park, William Chan, Yu~Zhang, Chung-Cheng Chiu, Barret Zoph, Ekin~D.
  Cubuk, and Quoc~V. Le,
\newblock ``{SpecAugment: A Simple Data Augmentation Method for Automatic
  Speech Recognition},''
\newblock in {\em Proc. Interspeech}, 2019, pp. 2613--2617.

\bibitem{Kingma14_iclr}
Diederik~P. {Kingma} and Jimmy {Ba},
\newblock ``{Adam: A Method for Stochastic Optimization},''
\newblock in {\em Proc. ICLR}, 2015.

\bibitem{Vaswani17_NIPS}
Ashish Vaswani, Noam Shazeer, Niki Parmar, Jakob Uszkoreit, Llion Jones,
  Aidan~N. Gomez, \L{}ukasz Kaiser, and Illia Polosukhin,
\newblock ``Attention is all you need,''
\newblock in {\em Proceedings of the 31st International Conference on Neural
  Information Processing Systems}, 2017, p. 6000–6010.

\bibitem{Chang21_asru}
Xuankai Chang, Takashi Maekaku, Pengcheng Guo, Jing Shi, Yen-Ju Lu,
  Aswin~Shanmugam Subramanian, Tianzi Wang, Shu-wen Yang, Yu~Tsao, Hung-yi Lee,
  and Shinji Watanabe,
\newblock ``{An Exploration of Self-Supervised Pretrained Representations for
  End-to-End Speech Recognition},''
\newblock in {\em Proc. ASRU}, 2021, pp. 228--235.

\bibitem{YYang22_arxiv}
Yuting Yang, Yuke Li, and Binbin Du,
\newblock ``Improving ctc-based asr models with gated interlayer
  collaboration,''
\newblock {\em ArXiv}, vol. abs/2205.12462, 2022.

\end{thebibliography}

\end{document}